\documentclass[runningheads]{llncs}
\usepackage[T1]{fontenc}
\usepackage{graphicx}
\usepackage{svg}
\usepackage{blindtext}
\usepackage{todonotes}
\usepackage{floatrow}
\newfloatcommand{capbtabbox}{table}[][\FBwidth]
\usepackage{moresize}

\definecolor{mygray}{gray}{0.9}
\definecolor{myorange}{RGB}{255, 178, 103}
\colorlet{light-orange}{myorange!20!white}

\usepackage{booktabs}
\usepackage{amsmath}
\usepackage{array,multirow}

\usepackage{hyperref}
\usepackage{color}

\begin{document}

\title{It's Just a Matter of Time: Detecting Depression with Time-Enriched Multimodal Transformers}
\titlerunning{A Time-Enriched Transformer for Multimodal Depression Detection}

%
\author{Ana-Maria Bucur\inst{1,2} \orcidID{0000-0003-2433-8877}\and
Adrian Cosma\inst{3} \orcidID{0000-0003-0307-2520}\and
Paolo Rosso\inst{2} \orcidID{0000-0002-8922-1242}\and
Liviu P. Dinu\inst{4} \orcidID{0000-0002-7559-6756}}
\authorrunning{Bucur et al.}


\institute{Interdisciplinary School of Doctoral Studies, University of Bucharest, Romania \and
PRHLT Research Center, Universitat Politècnica de València, Spain \and
Politehnica University of Bucharest, Romania \and
Faculty of Mathematics and Computer Science, University of Bucharest, Romania
\email{ana-maria.bucur@drd.unibuc.ro, cosma.i.adrian@gmail.com prosso@dsic.upv.es, ldinu@fmi.unibuc.ro
}}

\maketitle

\begin{abstract}
Depression detection from user-generated content on the internet has been a long-lasting topic of interest in the research community, providing valuable screening tools for psychologists. The ubiquitous use of social media platforms lays out the perfect avenue for exploring mental health manifestations in posts and interactions with other users. Current methods for depression detection from social media mainly focus on text processing, and only a few also utilize images posted by users. In this work, we propose a flexible time-enriched multimodal transformer architecture for detecting depression from social media posts, using pretrained models for extracting image and text embeddings. Our model operates directly at the user-level, and we enrich it with the relative time between posts by using time2vec positional embeddings. Moreover, we propose another model variant, which can operate on randomly sampled and unordered sets of posts to be more robust to dataset noise. We show that our method, using EmoBERTa and CLIP embeddings, surpasses other methods on two multimodal datasets, obtaining state-of-the-art results of 0.931 F1 score on a popular multimodal Twitter dataset, and 0.902 F1 score on the only multimodal Reddit dataset.

\end{abstract}

\section{Introduction}
More than half of the global population uses social media\footnote{\url{https://datareportal.com/reports/digital-2022-july-global-statshot}}. People use platforms such as Twitter and Reddit to disclose and discuss their mental health problems online. On Twitter, users feel a sense of community, it is a safe place for expression, and they use it to raise awareness and combat the stigma around mental illness or as a coping mechanism \cite{berry2017whywetweetmh}. On Reddit, more so than on Twitter, users are pseudo-anonymous, and they can choose to use "throw-away" accounts to be completely anonymous, subsequently encouraging users to disclose their mental health problems. Users talk about symptoms and their daily struggles, treatment, and therapy \cite{de2014mental} on dedicated subreddits such as \textit{r/depression, r/mentalhealth}. To date, there have been many methods that aim to estimate signs of mental disorders (i.e., depression, eating disorders) \cite{losada2019overview,yates2017depression} from the social media content of users. The primary focus has been on processing the posts' text, fuelled partly by the widespread availability and good performance of pretrained language models (e.g., BERT) \cite{alhuzali2021predicting,kenton2019bert,wu2021roberta}. Recently, however, both textual and visual information has been used for multimodal depression detection from social media data, on datasets collected from Twitter \cite{gui2019cooperative,safa2022automatic,shen2017depression}, Instagram \cite{chiu2021multimodal,mann2020see}, Reddit \cite{uban2022explainability}, Flickr \cite{xu2020inferring}, and Sina Weibo \cite{wang2020multimodal}. These methods obtain good performance, but nevertheless assume that social media posts are synchronous and uploaded at regular intervals.

We propose a time-enriched multimodal transformer for user-level depression detection from social media posts (i.e., posts with text and images). Instead of operating at the token-level in a low-level manner, we utilize the cross- and self-attention mechanism across posts to learn the high-level posting patterns of a particular user. The attention layers process semantic embeddings obtained by pretrained state-of-the-art language and image processing models. As opposed to current time-aware methods for mental disorders detection \cite{tlstm-original,cheng2022multimodal}, our method does not require major architectural modifications and can easily accommodate temporal information by simply manipulating the transformer positional encodings (e.g., using time-enriched encodings such as time2vec \cite{time2vec}). We propose two viable ways to train our architecture: a time-aware regime using time2vec and a set-based training regime, in which we do not employ positional encodings and regard the user posts as a set. The second approach is motivated by the work of Dufter et al. \cite{dufter2021position}, which observed that positional encodings are not universally necessary to obtain good downstream performance. We train and evaluate our method on two social media multimodal datasets, each with its own particularities in user posting behavior, and obtain state-of-the-art results in depression detection. We make our code publicly available on github\footnote{\url{https://github.com/cosmaadrian/time-enriched-multimodal-depression-detection}}.

This work makes the following contributions:
\begin{enumerate}
    \item We propose a time-enriched multimodal transformer for user-level depression detection from social media posts. Using EmoBERTa and CLIP embeddings, and time2vec positional embeddings, our method achieves 0.931 F1 on a popular multimodal Twitter dataset \cite{gui2019cooperative}, surpassing current methods by a margin of 2.3\%. Moreover, using no positional embeddings, we achieve  0.902 F1 score on \textit{multiRedditDep} \cite{uban2022explainability}, the only multimodal Reddit dataset to date.
    
    \item We perform extensive ablation studies and evaluate different types of image and text encoders, window sizes, and positional encodings. We show that a time-aware approach is suitable when posting frequency is high, while a set-based approach is robust to dataset noise (i.e., many uninformative posts).
    
    \item We perform a qualitative error analysis using Integrated Gradients \cite{sundararajan2017axiomatic}, which proves that our model is interpretable and allows for the selection of the most informative posts in a user's social media timeline.
    
\end{enumerate}

\section{Related Work}
Although research in depression detection was focused on analyzing language cues uncovered from psychology literature (e.g., self-focused language reflected in the greater use of the pronoun "I" \cite{rissola2020beyond,rude2004language}, dichotomous thinking expressed in absolute words (e.g., "always", "never") \cite{fekete2002internet}), studies also began to investigate images posted on social media. Reece et al. \cite{reece2017instagram} showed that the images posted online by people with depression were more likely to be sadder and less happy, and to have bluer, darker and grayer tones than those from healthy individuals. Users with depression posted more images with faces of people, but they had fewer faces per image, indicating reduced social interactivity and an increased self-focus \cite{guntuku2019twitter,reece2017instagram}. Guntuku et al. \cite{guntuku2019twitter} and Uban et al. 
\cite{uban2022explainability} revealed that users diagnosed with depression have more posts with animal-related images. 

Deep learning methods such as CNNs \cite{rao2020mgl,yates2017depression}, LSTMs \cite{skaik2020using,trotzek2018utilizing} and transformer-based architectures \cite{alhuzali2021predicting,bucur2021early,wu2021roberta} achieved good results on depression detection using only the textual information from users' posts. Further, multimodal methods incorporating visual features achieve even greater results \cite{mann2020see,safa2022automatic,xu2020inferring}. Shen et al. \cite{shen2017depression} collected the first user-level multimodal dataset from social media for identifying depression with textual, behavioral, and visual information from Twitter users and proposed a multimodal depressive dictionary learning method. The same Twitter dataset was later explored by Gui et al. \cite{gui2019cooperative}, who used a cooperative multi-agent reinforcement learning method with two agents for selecting only the relevant textual and visual information for classification. An et al. \cite{an2020multimodal} proposed a multimodal topic-enriched auxiliary learning approach, in which the performance of the primary task on depression detection is improved by auxiliary tasks on visual and textual topic modeling. By not taking the time component into account, the above methods assume that social media posts are synchronous, and are sampled at regular time intervals. Realistically, posts from online platforms are asynchronous and previous studies have shown differences in social media activity, partly due to the worsening of depression symptoms at night \cite{de2013predicting,lustberg2000depression,stankevich2018feature}. Motivated by this, methods that use time-adapted weights \cite{chiu2021multimodal} or Time-Aware LSTMs \cite{cheng2022multimodal,sawhney2020time} to include the time component of data for mental health problems detection report higher performance.

\section{Method}
 \begin{figure*}[hbt!]
    \centering
    \includegraphics[width=1.0\textwidth]{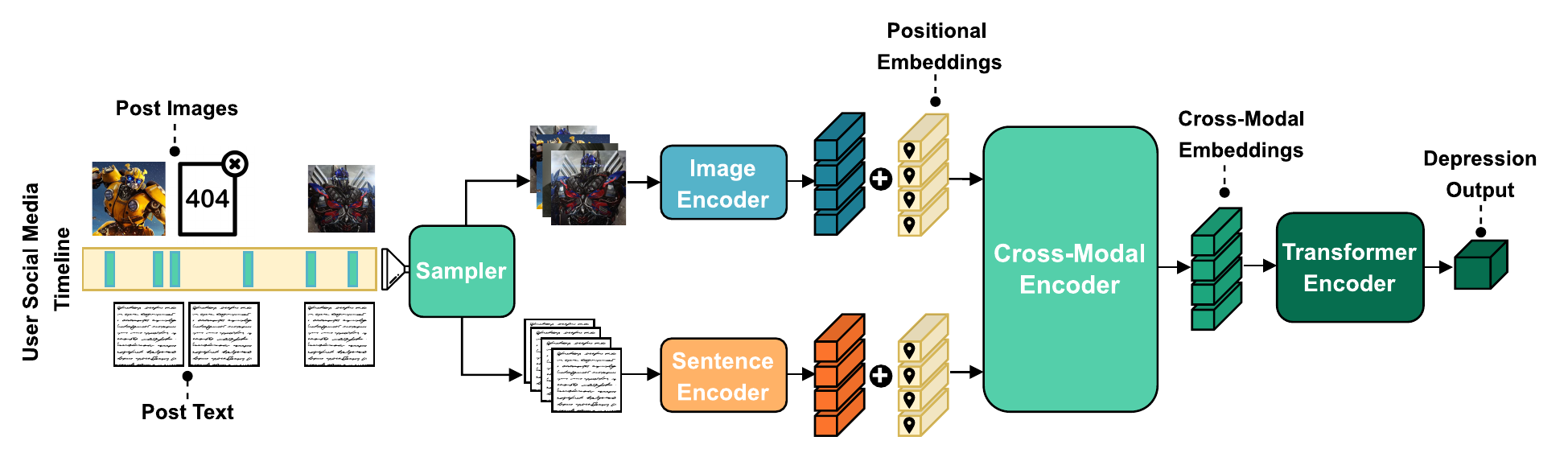}
    \caption{The overall architecture of our proposed method. From a user's social media timeline, we sample a number of posts containing text and images. Each image and text is encoded with pretrained image and text encoders. The sequence of encoded images and texts is then processed using a cross-modal encoder followed by a transformer encoder, together with the relative posting time encoded into the positional embeddings. After mean pooling, we perform classification for the particular user.}
    \label{fig:diagram}
\end{figure*}

The problem of depression detection from social media data is usually formulated as follows: given a user with an ordered, asynchronous, sequence of multimodal social media posts containing text and images, determine whether or not the user has symptoms of depression. This formulation corresponds to user-level binary classification. The main difficulty in this area is modeling a large number of user posts (i.e., tens of thousands), in which not all posts contain relevant information for depression detection. This aspect makes the classification inherently noisy. Formally, we consider a user $i$ to have multiple social media posts $U_i$, each post containing the posting date $\Delta$, a text $T$ and an accompanying image $I$. A post-sequence $P_i$ is defined as $K$ posts sampled from $U_i$: $P_i = \{(T^j, I^j, \Delta^j) \sim U_i, j \in (1\dots K)\}$. During training, we used an input batch defined by the concatenation of $n$ such post-sequences: $B = \{S_{b_1}, S_{b_2}, \dots S_{b_n}\}$. Since the users' posts are asynchronous (i.e., are not regularly spaced in time), time-aware approaches based on T-LSTM \cite{cheng2022multimodal,sawhney2020time} have become the norm in modeling users' posts alongside with the relative time between them. However, in T-LSTM \cite{tlstm-original}, including a relative time component involves the addition of new gates and hand-crafted feature engineering of the time. Moreover, T-LSTM networks are slow, cannot be parallelized and do not allow for transfer learning.

To address this problem, we propose a transformer architecture that can perform user-level multimodal classification, as shown in Figure \ref{fig:diagram}. To process the multimodal data from posts, we first encode the visual and textual information using pretrained models (e.g., CLIP \cite{radford2021learning} for images and EmoBERTa \cite{kim2021emoberta} for text). The embeddings are linearly projected to a fixed size using a learnable feed-forward layer, are augmented with a variant of positional encodings (expanded below) and are further processed with a cross-modal encoder based on LXMERT \cite{tan2019lxmert}. Finally, self-attention is used to process the cross-modal embeddings further, and classification is performed after mean pooling. The network is trained using the standard binary cross-entropy loss. In this setting, the transformer attention does not operate on low-level tokens such as word pieces or image patches. Rather, the cross- and self-attention operates across posts, allowing the network to learn high-level posting patterns of the user, and be more robust to individual uninformative posts. As opposed to other related works in which a vector of zeros replaces the embeddings of missing images \cite{an2020multimodal,gui2019cooperative}, we take advantage of the attention masking mechanism present in the transformer architecture \cite{vaswani2017attention}, and mask out the missing images.

\begin{figure}[hbt!]
    \centering
    \includegraphics[width=0.55\linewidth]{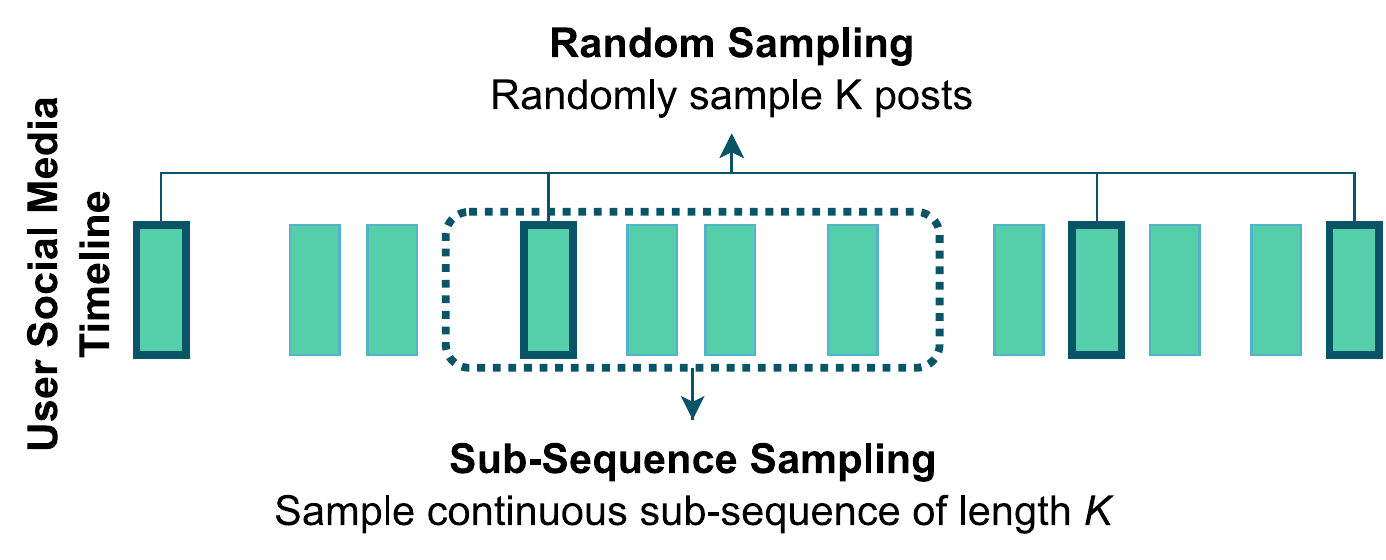}
    \caption{The two sampling methods used in this work. Methods that use positional embeddings (either \textit{time2vec} or \textit{learned}) employ sub-sequence sampling, and the \textit{SetTransformer}, with \textit{zero} positional embeddings, uses random sampling of posts. $K$ refers to the number of posts in the post window.}
    \label{fig:sampling}
\end{figure}

For this work, we experiment with positional embeddings based on \textit{time2vec} \cite{time2vec} to make the architecture time-aware. Utilizing \textit{time2vec} as a method to encode the time $\tau$ into a vector representation is a natural way to inject temporal information without any major architectural modification, as it was the case for T-LSTM, for example. \textit{Time2vec} has the advantage of being invariant to time rescaling, avoiding hand-crafted time features, it is periodic, and simple to define and consume by various models. It is a vector of $k + 1$ elements, defined as:

\begin{equation}
    \text{\textbf{t2v}}(\tau)[i] = 
    \left\{
	\begin{array}{ll}
		\omega_i g(\tau) + \phi_i  &  i = 0 \\
		\mathcal{F}(\omega_ig(\tau) + \phi_i) & 1 \leq i \leq k
	\end{array}
\right.
\end{equation}
where \textbf{t2v}($\tau$)[i] is the $i^{\text{th}}$ element of \textbf{t2v}($\tau$), $\mathcal{F}$ is a periodic activation function (in our case it is $sin(x)$), and $\omega_is$ and $\phi_is$ are learnable parameters. To avoid arbitrarily large $\tau$ values, we transform $\tau$ with $g(\tau) = \frac{1}{(\tau + \epsilon)}$, with $\epsilon = 1$. For processing user's posts, we use sub-sequence sampling and sample $K$ \textit{consecutive} posts from a user's timeline (Figure \ref{fig:sampling}). 
We name the model utilizing \textit{time2vec} embeddings \textit{Time2VecTransformer}.

However, processing user posts sequentially is often not desirable, as clusters of posts might provide irrelevant information for depression detection. For instance, sub-sequence sampling of posts from users with mental health problems might end up in an interval of positive affect, corresponding to a sudden shift in mood \cite{tsakalidis2022identifying}. Somewhat orthogonal to the time-aware school of thought, we also propose a \textit{SetTransformer} for processing sets of user posts for depression detection, to alleviate the issues mentioned above. Our proposed \textit{SetTransformer} randomly samples texts from a user and assumes no order between them by omitting the positional encoding, essentially making the transformer permutation invariant \cite{vaswani2017attention}. For this method, K posts in a post-sequence are randomly sampled (Figure \ref{fig:sampling}). For \textit{SetTransformer}, we treat the user timeline as a "bag-of-posts" motivated by the work of Dufter et al. \cite{dufter2021position}, in which the authors show that treating sentences as "bag-of-words" (i.e., by not utilizing any positional embeddings) results in a marginal loss in performance for transformer architectures.

\section{Experiments}
\subsection{Datasets}

To benchmark our method, we use in our experiments \textit{multiRedditDep} \cite{uban2022explainability} and the Twitter multimodal depression dataset from Gui et al. \cite{gui2019cooperative}\footnote{We also attempted to perform our experiments on a multimodal dataset gathered from Instagram \cite{cheng2022multimodal,chiu2021multimodal}, but the authors did not respond to our request.}. Reddit and Twitter are two of the most popular social media platforms where users choose to disclose their mental health problems \cite{berry2017whywetweetmh,de2014mental}. Moreover, the data coming from these two platforms have different particularities: both social media platforms support images, but the textual information is richer on Reddit, with posts having up to 40,000 characters, as opposed to Twitter, where the limit is 280 characters. 
In some subreddits from Reddit, the image cannot be accompanied by text; the post is composed only of image(s) and a title with a maximum length of 300 characters. On Twitter, posts with images have the same character limit as regular posts. For both datasets, the users from the depression class were annotated by retrieving their mention of diagnosis, while users from the control group did not have any indication of depression diagnosis. 

\begin{figure}
\begin{floatrow}
\ffigbox{%
    \includegraphics[width=1.0\linewidth]{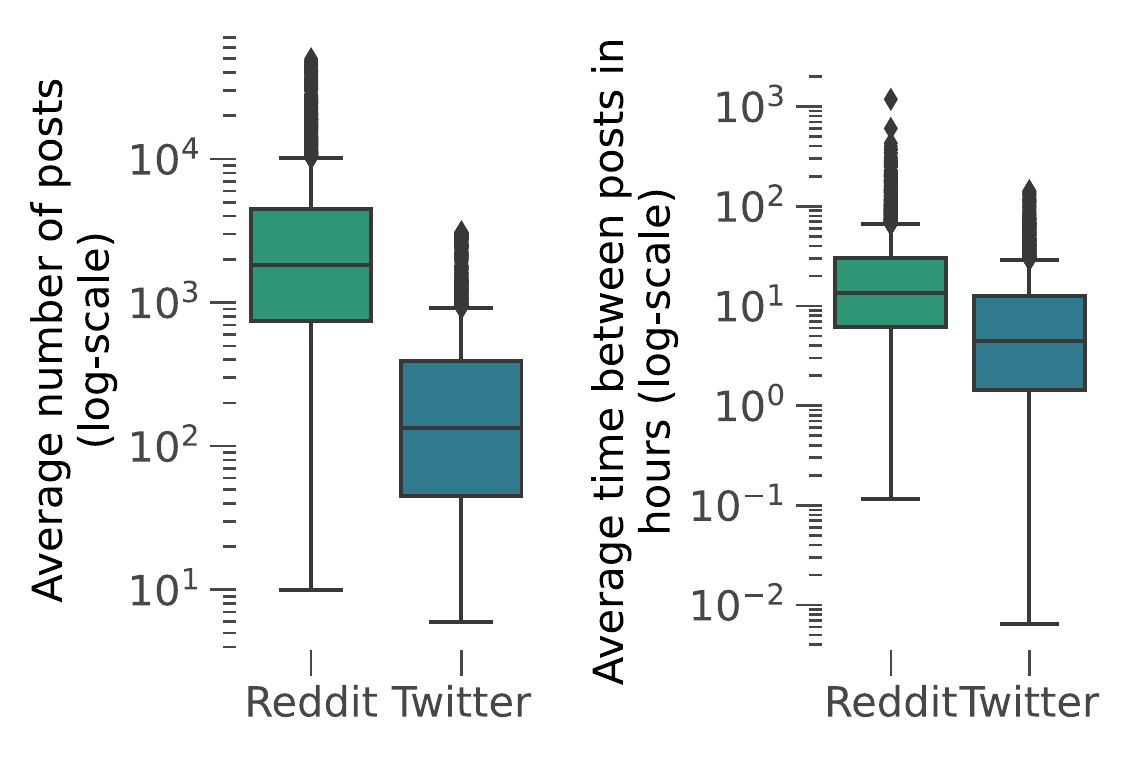}
}{%
    \caption{\textit{Left} - Distribution of posts per user on both datasets. Users from Reddit have significantly (Mann Whitney U Test, p < 0.001) more posts than users from Twitter. \textit{Right} - Distribution of average time duration in hours between posts at the user level. Users from Twitter post significantly (p < 0.001) more frequently than users from Reddit.}
    \label{fig:post-duration}
}
\capbtabbox{%
    \resizebox{\linewidth}{!}{
  \begin{tabular}{c|rcccc}
        \textbf{Dataset} & \textbf{Class} & \#T & \#(T+I) & $\overline{\text{\#T}}$ & $\overline{\text{\#(T+I)}}$\\
        \toprule
        \multirow{2}{*}{Reddit} & Depr & 6,6M & 46.9k & 4.7k & 33.33\\
         & Non-Depr & 8,1M & 73.4k & 3.5k & 31.51\\
        \midrule
        \multirow{2}{*}{Twitter} & Depr & 213k & 19.3k & 152.43 & 13.81 \\
         & Non-Depr & 828k & 50.6k & 590.82 & 36.16\\
        \end{tabular}
        }
}{%
 \caption{Statistics for the Reddit \cite{uban2022explainability} and Twitter \cite{gui2019cooperative} datasets. \#T represents the number of posts with only text, \#(T+I) represents the number of posts with both text (title, in the case of Reddit) and images. $\overline{\text{\#T}}$ and $\overline{\text{\#(T+I)}}$ represent the average number of posts with only text and text + image per user, respectively.}
 \label{tab:datasets-statistics}
}
\end{floatrow}
\end{figure}

The Reddit dataset contains 1,419 users from the depression class and 2,344 control users. The authors provided the train, validation and test splits, with 2633, 379 and 751 users, respectively. The dataset from Twitter contains 1,402 users diagnosed with depression and 1,402 control users. Shen et al. \cite{shen2017depression} collected only tweets in the span of one month for users from both classes. Due to the lack of access to the benchmark train/test splits from An et al. \cite{an2020multimodal}, we are running the experiments following the same experimental settings as Gui et al. \cite{gui2019cooperative}, and perform five-fold cross-validation. In Table \ref{tab:datasets-statistics}, we showcase the statistics for both datasets. There is a greater total number of posts from the non-depressed group, but depressed users have more posts, on average, than non-depressed users, in the case of Reddit. For Twitter, the opposite is true, with non-depressed users having more posts on average. It is important to note that not all posts have images, only a small amount of social media posts contain both text and image data. Figure \ref{fig:post-duration} (left) shows that Reddit users have a greater average number of posts than Twitter users. Regarding posting frequency, Twitter users post more frequently than Reddit users, as shown in Figure \ref{fig:post-duration} (right), in which the average time between posts is 26.9 hours for Reddit and 11.2 hours for Twitter.


\subsection{Experimental Settings}

\textbf{Image representation methods.}
For encoding the images in users' posts, we opted to use two different encoders trained in a self-supervised manner. Many approaches for transfer learning employ a supervised network, usually pretrained on ImageNet \cite{deng2009imagenet}. However, transfer learning with self-supervised trained networks is considered to have more general-purpose embeddings \cite{caron2021emerging}, which aid downstream performance, especially when the training and transfer domains qualitatively differ \cite{ericsson2021well}. Since images in our social media datasets are very diverse, including internet memes and screenshots \cite{uban2022explainability} with both visual and textual information, specialized embeddings from a supervised ImageNet model are not appropriate. Therefore, we used \textbf{CLIP} \cite{radford2021learning}, a vision transformer trained with a cross-modal contrastive objective, to connect images with textual descriptions. CLIP has been shown to perform well on a wide range of zero-shot classification tasks, and is capable of encoding text present in images. CLIP embeddings are general, multi-purpose, and are trained on a large-scale internet-scraped dataset. Additionally, we used \textbf{DINO} \cite{caron2021emerging}, a vision transformer pretrained in a self-supervised way on ImageNet, achieving notable downstream performance. Moreover, compared to a supervised counterpart, DINO automatically learns class-specific features without explicitly being trained to do so. 

\textbf{Text representation methods.} We explored three pretrained transformers models to extract contextual embeddings from users' texts. \textbf{RoBERTa} \cite{liu2019roberta}, pretrained in a self-supervised fashion on a large corpus of English text, was chosen given its state-of-the-art performance on downstream tasks. Emotion-informed embeddings from \textbf{EmoBERTa} \cite{kim2021emoberta} that incorporate both linguistic and emotional information from users' posts were also used. Emotions expressed in texts are a core feature used for identifying depression \cite{aragon2021detecting,lara2021deep}, users with depression showing greater negative emotions \cite{de2013predicting}. EmoBERTa is based on a RoBERTa model trained on two conversational datasets and adapted for identifying the emotions found in users' utterances.  \textbf{Multilingual MiniLM} \cite{wang2020minilm}, the distilled version of the multilingual language model XLM-RoBERTa \cite{conneau-etal-2020-unsupervised}, was used for encoding text because both Twitter and Reddit datasets contain posts in various languages besides English, such as Spanish, German, Norwegian, Japanese, Romanian and others. Moreover, MiniLM is a smaller model, providing 384-dimensional embeddings, as opposed to 768-dimensional embeddings from RoBERTa and EmoBERTa.

\textbf{Positional encodings.}
Since the network is processing sequences of posts, we explored three different methods for encoding the relative order between posts. Firstly, we used a standard \textbf{learned} positional encoding, used in many pretrained transformer models, such as BERT \cite{kenton2019bert}, RoBERTa \cite{liu2019roberta} and GPT \cite{brown2020language}. Wang and Chen \cite{wang-chen-2020-position} showed that for transformer encoder models, learned positional embeddings encode local position information, which is effective in masked language modeling. However, this type of positional encoding assumes that users' posts are equally spaced in time. Second, we used \textbf{time2vec} \cite{time2vec} positional encoding, which allows the network to learn a vectorized representation of the relative time between posts. Lastly, we omit to use any positional encoding, and treat the sequence of user posts as a set. We refer to this type of positional encodings as \textbf{zero} encodings. For both \textit{learned} and \textit{time2vec} we used a sub-sequence sampling of posts, while for \textit{zero}, we used a random sample of user posts (Figure \ref{fig:sampling}).


\subsection{Comparison Models}
We evaluate our proposed method's performance against existing multimodal and text-only state-of-the-art models on the two multimodal datasets from Twitter and Reddit, each with its different particularities in user posting behavior. For \textit{multiRedditDep}, since it is a new dataset, there are no public benchmarks besides the results of Uban et al. \cite{uban2022explainability}. We report Accuracy, Precision, Recall, F1, and AUC as performance measures. The baselines are as follows. \textbf{Time-Aware LSTM (T-LSTM)} \cite{tlstm-original} - we implement as text-only baseline a widely used \cite{cheng2022multimodal,sawhney2020time} T-LSTM-based neural network architecture that integrates the time irregularities of sequential data in the memory unit of the LSTM. \textbf{EmoBERTa Transformer} - we train a text-only transformer baseline on user posts. Both text-only models are based on EmoBERTa embeddings. \textbf{LSTM + RL} and \textbf{CNN + RL} \cite{gui2019depression} - two text-only state-of-the-art models which use a reinforcement learning component for selecting the posts indicative of depression. \textbf{Multimodal Topic-Enriched Auxiliary Learning (MTAL)} \cite{an2020multimodal} - a model capturing the multimodal topic information, in which two auxiliary tasks accompany the primary task of depression detection on visual and textual topic modeling. \textbf{Multimodal Time-Aware Attention Networks (MTAN)} \cite{cheng2022multimodal} - a multimodal model that uses as input BERT \cite{kenton2019bert} textual features, InceptionResNetV2 \cite{Szegedy2017inception} visual features, posting time features and incorporates T-LSTM for taking into account the time intervals between posts and self-attention. \textbf{GRU + VGG-Net + COMMA} \cite{gui2019cooperative} - in which a reinforcement learning component is used for selecting posts with text and images which are indicative of depression and are classified with an MLP. For extracting the textual and visual features, GRU \cite{chung2014empirical} and VGGNet \cite{DBLP:journals/corr/SimonyanZ14a} were used. \textbf{BERT + word2vec embeddings} - baseline proposed by Uban et al. \cite{uban2022explainability}, which consists of a neural network architecture that uses as input BERT features from posts' titles and word2vec embeddings for textual information found in images (i.e., ImageNet labels and text extracted from images). \textbf{VanillaTransformer} - the multimodal transformer proposed in this work, with standard \textit{learned} positional encodings. \textbf{SetTransformer} - the set-based multimodal transformer proposed in this work employing \textit{zero} positional encoding, alongside a random sampling of user posts. \textbf{Time2VecTransformer} - the time-aware multimodal transformer proposed in this work using time-enriched positional embeddings (i.e., \textit{time2vec} \cite{time2vec}) and sub-sequence sampling.

\subsection{Training and Evaluation Details}

We train all models using Adam \cite{kingma2014method} optimizer with a base learning rate of 0.00001. The learning rate is modified using Cyclical Learning Rate schedule \cite{DBLP:journals/corr/Smith15a}, which linearly varies the learning rate from 0.00001 to 0.0001 and back across 10 epochs. The model has 4 cross-encoder layers with 8 heads each and an embedding size of 128. The self-attention transformer has 2 layers of 8 heads each and the same embedding size. 

At test time, since it is unfeasible to use all users' posts for evaluation, with some users having more than 50k posts, we make 10 random samples of post-sequences for a user, and the final decision is taken through majority voting on the decisions for each post-sequence. In this way, the final classification is more robust to dataset noise and uninformative posts.

\section{Results}
\subsection{Performance Comparison with Prior Works}

In Table \ref{tab:twitter-sota}, we present the results for models trained on the Twitter dataset. For our models and proposed baselines, each model was evaluated 10 times and we report mean and standard deviation. Our architecture, \textit{Time2VecTransformer} achieves state-of-the-art performance in multimodal depression detection, obtaining a 0.931 F1 score. The model uses as input textual embeddings extracted from EmoBERTa and visual embeddings extracted from CLIP. The best model uses sequential posts as input from a 512 window size, and the time component is modeled by \textit{time2vec} positional embeddings. Our time-aware multimodal architecture surpasses other time-aware models such as T-LSTM \cite{tlstm-original} and MTAN \cite{cheng2022multimodal}. Moreover, our method surpasses text-only methods such as T-LSTM and EmoBERTa Transformer - a unimodal variant of our model using only self-attention on text embeddings.

\begin{table*}[hbt!]
    \caption{Results for multimodal depression detection on Twitter dataset \cite{gui2019cooperative}. Our models use EmoBERTa and CLIP for extracting embeddings. \textit{VanillaTransformer} and \textit{SetTransformer} were trained on 128 sampled posts, while \textit{Time2VecTransformer} was trained on a window size of 512 posts. Denoted with \textbf{bold} are the best results for each column. $^*$An et al. \cite{an2020multimodal} report the performance on a private train/dev/test split, not on five-fold cross-validation. $^{**}$ Cheng et al. \cite{cheng2022multimodal} are not explicit in their experimental settings for the Twitter data. $^\dagger$ indicates that the result is a statistically significant improvement over SetTransformer ($p<0.005$, using Wilcoxon signed-rank test). $^\ddagger$ indicates that there is a statistically significant improvement over Time2VecTransformer ($p<0.05$, using Wilcoxon signed-rank test).}
    \centering
    \resizebox{\linewidth}{!}{
    \begin{tabular}{l|c|cccc}
         \textbf{Method} & \textbf{Modality} & \textbf{F1} & \textbf{Prec.} & \textbf{Recall} & \textbf{Acc.}\\
         \toprule
         T-LSTM \cite{tlstm-original} & T & 0.848{\ssmall$\pm$8e-3} & 0.896{\ssmall$\pm$2e-2} & 0.804{\ssmall$\pm$1e-2} & 0.855{\ssmall$\pm$5e-3} \\
         EmoBERTa Transformer & T & 0.864{\ssmall$\pm$1e-2} & 0.843{\ssmall$\pm$1e-2} & 0.887{\ssmall$\pm$3e-2} & 0.861{\ssmall$\pm$1e-2} \\
         LSTM + RL \cite{gui2019depression} & T & 0.871 & 0.872 & 0.870 & 0.870\\
         CNN + RL \cite{gui2019depression} & T & 0.871 & 0.871 & 0.871 & 0.871\\
         \midrule
         MTAL \cite{an2020multimodal}$^\ast$ & T+I & 0.842 & 0.842 & 0.842 & 0.842\\
         GRU + VGG-Net + COMMA \cite{gui2019cooperative} & T+I & 0.900 & 0.900 & 0.901 & 0.900\\
         MTAN \cite{cheng2022multimodal}$^{**}$ & T+I & 0.908 & 0.885 & 0.931 & -\\
         \midrule
         Vanilla Transformer \textbf{(ours)} & T+I & 0.886{\ssmall$\pm$1e-2} & 0.868{\ssmall$\pm$2e-2} &  0.905{\ssmall$\pm$2e-2} & 0.883{\ssmall$\pm$5e-3} \\
         SetTransformer  \textbf{(ours)} & T+I & 0.927{\ssmall$\pm$8e-3} & 0.921{\ssmall$\pm$1e-2} &  \textbf{0.934{\ssmall$\pm$2e-2}$^\ddagger$} & 0.926{\ssmall$\pm$8e-3} \\
         Time2VecTransformer  \textbf{(ours)} & T+I & \textbf{0.931{\ssmall$\pm$4e-3}$^\dagger$} &	\textbf{0.931{\ssmall$\pm$2e-2}$^\dagger$} & 0.931{\ssmall$\pm$1e-2} &	\textbf{0.931{\ssmall$\pm$4e-3}$^\dagger$} \\ 
    \end{tabular}
    }
    \label{tab:twitter-sota}
\end{table*}
\begin{table*}[hbt!]
    \centering
    \caption{Results for multimodal depression detection on \textit{multiRedditDep}. Our models use EmoBERTa and CLIP for extracting embeddings. \textit{VanillaTransformer} was trained on 128 sampled posts, while \textit{Time2VecTransformer} and \textit{SetTransformer} were trained on a window size of 512 posts. We denote with \textbf{bold} the best results for each column. $^*$Uban et al. \cite{uban2022explainability} conducted experiments using the visual and textual features from images and titles of the posts. $^\dagger$ indicates that the result is a statistically significant improvement over Time2VecTransformer ($p<0.005$, using Wilcoxon signed-rank test).}
    \resizebox{\linewidth}{!}{
    \begin{tabular}{l|c|ccccc}
        \textbf{Method} & \textbf{Modality} & \textbf{F1} & \textbf{Prec.} & \textbf{Recall} & \textbf{Acc.} & \textbf{AUC}\\
        \toprule
        T-LSTM \cite{tlstm-original} & T & 0.831{\ssmall$\pm$1e-2} & 0.825{\ssmall$\pm$8e-3} & 0.837{\ssmall$\pm$1e-2} & 0.872{\ssmall$\pm$7e-3} & 0.946{\ssmall$\pm$2e-3} \\
        EmoBERTa Transformer & T & 0.843{\ssmall$\pm$6e-3} & 0.828{\ssmall$\pm$3e-3} & 0.858{\ssmall$\pm$1e-2} & 0.879{\ssmall$\pm$4e-3} & 0.952{\ssmall$\pm$2e-3} \\
        \midrule
        Uban et al. \cite{uban2022explainability}$^*$ & T+I & - & - & - & 0.663 & 0.693\\
        \midrule
        VanillaTransformer \textbf{(ours)} & T+I & 0.837{\ssmall$\pm$8e-3} & 0.827{\ssmall$\pm$1e-2} & 0.848{\ssmall$\pm$1e-2} & 0.876{\ssmall$\pm$6e-3} & 0.956{\ssmall$\pm$3e-3} \\
        SetTransformer \textbf{(ours)} & T+I & \textbf{0.902{\ssmall$\pm$7e-3}$^\dagger$} & \textbf{0.878{\ssmall$\pm$6e-3}$^\dagger$} &  \textbf{0.928{\ssmall$\pm$1e-2}$^\dagger$} & \textbf{0.924{\ssmall$\pm$5e-3}$^\dagger$} & \textbf{0.976{\ssmall$\pm$1e-3}$^\dagger$}  \\
        Time2VecTransformer \textbf{(ours)} & T+I & 0.869{\ssmall$\pm$7e-3} & 0.869{\ssmall$\pm$7e-3} & 0.869{\ssmall$\pm$8e-3} & 0.901{\ssmall$\pm$5e-3} & 0.967{\ssmall$\pm$1e-3} \\
    \end{tabular}
    }
    \label{tab:reddit-sota}

\end{table*}

As opposed to the previous result on Twitter data, which include the time component, for Reddit, we achieved good performance by regarding the users' posts as a set. In Table \ref{tab:reddit-sota}, we showcase our model's performance on the Reddit dataset, compared to Uban et al. \cite{uban2022explainability}, our trained text-only T-LSTM and text-only EmoBERTa Transformer. Our \textit{SetTransformer} model (with \textit{zero} positional encodings), using as input EmoBERTa and CLIP, obtains the best performance, a 0.902 F1 score. While counter-intuitive, treating posts as a "bag-of-posts" outperforms \textit{Time2VecTransformer} in the case of Reddit. However, the Reddit dataset contains a considerable amount of noise and uninformative posts (links, short comments, etc.), which dilutes discriminative information for depression detection. A random sampling of posts seems to alleviate this problem to some degree.

\subsection{Ablation Study}

\begin{figure}[hbt!]
    \includegraphics[width=0.8\linewidth]{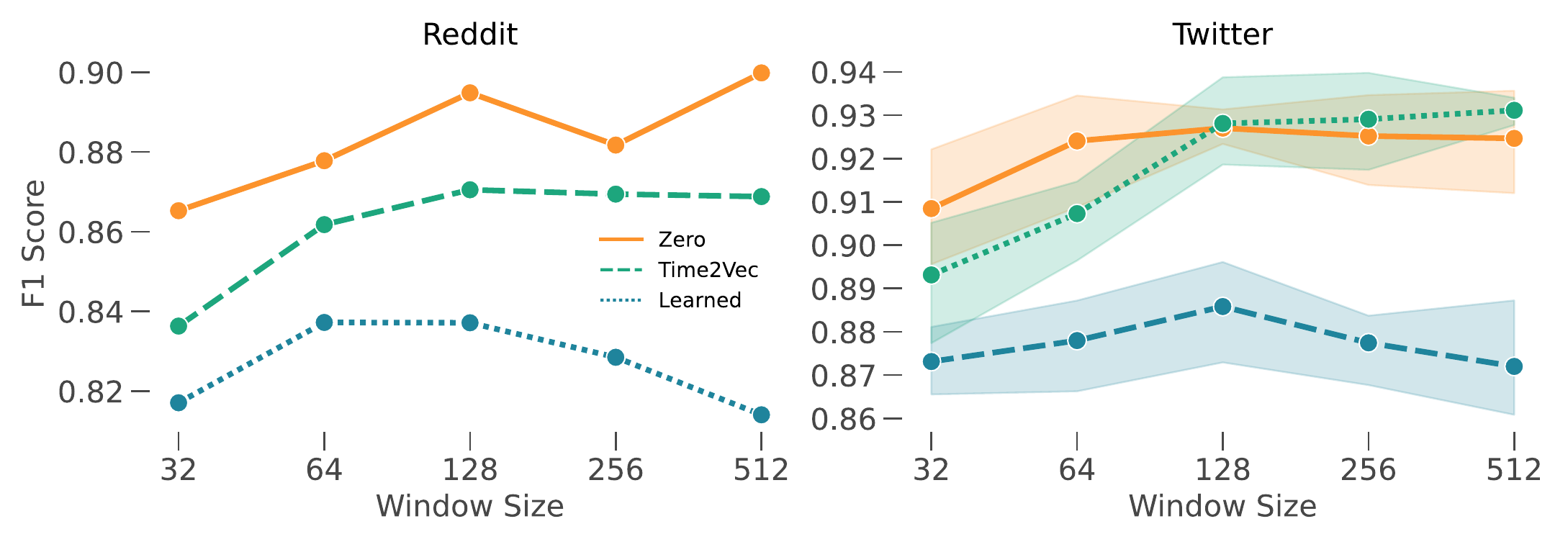}{
    \caption{Comparison among different window sizes used in our experiments, using EmoBERTa for text embeddings and CLIP for image encodings. For Twitter, the results were averaged across the 5 folds.}
    \label{fig:windowsize-comparison}
}
\end{figure}

To gauge the effect of the sampling window size, we performed experiments using CLIP as an image encoder and EmoBERTa as a text encoder, in which we varied the window size in $K = \{32, 64, 128, 256, 512\}$, as presented in Figure \ref{fig:windowsize-comparison}. For Reddit, the 128 window size is the best suited for all three kinds of positional embeddings, as evidenced by the high F1 score. Even if Reddit users have an average of over 3,000 posts (Table \ref{tab:datasets-statistics}), 128 posts contain enough information to make a correct decision. On Twitter, for \textit{learned} and \textit{zero} positional embeddings, the model with 128 posts window size performs best, while for \textit{time2vec}, 512 has the best F1 score. This may be because the 512 window size covers the average number of posts from users in the Twitter dataset (see Table \ref{tab:datasets-statistics}). 
Given the short time span of one month for the posts coming from Twitter, we can hypothesize that for datasets in which the time between posts is very small (average of 11.2 hours for Twitter), the time component modeled by \textit{time2vec} positional encodings may be more informative than other positional embedding methods. For Reddit, many of the users' posts are, in fact, comments or links, which are usually not informative to the model decision. Nevertheless, we achieve a performance comparable to the previous state-of-the-art, even in low-resource settings, by processing only 32 posts. Including the time component modeled by \textit{time2vec} has a more important contribution when the time between posts is shorter (as in the data from Twitter) as opposed to larger periods between the posts (as is the case of Reddit).


In Table \ref{tab:reddit-ablation}, we showcase different combinations of text encoders (RoBERTa / EmoBERTa / MiniLM) and image encoders (CLIP / DINO). The difference between image encoders is marginal, due to the small number of images in the users' timeline (Table \ref{tab:datasets-statistics}). Interestingly, RoBERTa embeddings are more appropriate for Twitter, while EmoBERTa is better suited for modeling posts from Reddit. We hypothesize that this is due to the pseudo-anonymity offered through Reddit, encouraging users to post more intimate and emotional texts \cite{de2014mental}. Using RoBERTa text embeddings and CLIP image embeddings, we obtain an F1 score of 0.943, which is even higher than the state-of-the-art. Further, the performance using MiniLM for text embeddings is lacking behind other encoders.

\begin{table}[hbt!]
    \centering
    \caption{Model comparison with different text and image encoding methods using the Reddit and Twitter datasets, with \textit{time2vec} positional embeddings and 128 window size. The best results are with \textbf{bold}, and with \underline{underline} the second best results.}
    \resizebox{\linewidth}{!}{
    \begin{tabular}{l|ccc|ccc}
        & \multicolumn{3}{c}{\textbf{Reddit}} & \multicolumn{3}{c}{\textbf{Twitter}}\\
        
        \textbf{Text+Image Enc.} & \textbf{F1} & \textbf{Prec.} & \textbf{Recall} & \textbf{F1} & \textbf{Prec.} & \textbf{Recall}\\
        \toprule
        MiniLM+CLIP & 0.789{\ssmall$\pm$6e-3} & 0.686{\ssmall$\pm$7e-3} & \textbf{0.929{\ssmall$\pm$9e-3}} & 0.827{\ssmall$\pm$8e-3} & 0.803{\ssmall$\pm$3e-2} & 0.854{\ssmall$\pm$4e-2} \\
        MiniLM+DINO & 0.799{\ssmall$\pm$9e-3} & 0.745{\ssmall$\pm$8e-3} & \underline{0.862{\ssmall$\pm$1e-2}} & 0.792{\ssmall$\pm$8e-3} & 0.782{\ssmall$\pm$3e-2}  & 0.806{\ssmall$\pm$4e-2} \\
        \midrule
        RoBERTa+CLIP & 0.845{\ssmall$\pm$8e-3} & 0.829{\ssmall$\pm$8e-3} & \underline{0.862{\ssmall$\pm$1e-2}} & \textbf{0.943{\ssmall$\pm$6e-3}}	 & \textbf{0.951{\ssmall$\pm$1e-2}} & \textbf{0.936{\ssmall$\pm$2e-2}} \\
        RoBERTa+DINO  & 0.840{\ssmall$\pm$9e-3} & 0.820{\ssmall$\pm$7e-3} & 0.861{\ssmall$\pm$1e-2} & \underline{0.936{\ssmall$\pm$1e-2}} &  \underline{0.946{\ssmall$\pm$2e-2}} & \underline{0.926{\ssmall$\pm$2e-2}} \\
        \midrule
        EmoBERTa+CLIP  & \textbf{0.871{\ssmall$\pm$7e-3}} & \textbf{0.883{\ssmall$\pm$8e-3}} & 0.858{\ssmall$\pm$8e-3} & 0.928{\ssmall$\pm$1e-2} & 0.933{\ssmall$\pm$1e-2}  & 0.924{\ssmall$\pm$2e-2}  \\
        EmoBERTa+DINO  & \underline{0.863{\ssmall$\pm$6e-3}} & \underline{0.865{\ssmall$\pm$4e-3}} & \underline{0.862{\ssmall$\pm$1e-2}} & 0.915{\ssmall$\pm$1e-2} & 0.918{\ssmall$\pm$2e-2} & 0.913{\ssmall$\pm$2e-2} \\
    \end{tabular}
    }
    \label{tab:reddit-ablation}
\end{table}

\subsection{Error Analysis}

\begin{figure}[hbt!]
    \includegraphics[width=1\linewidth]{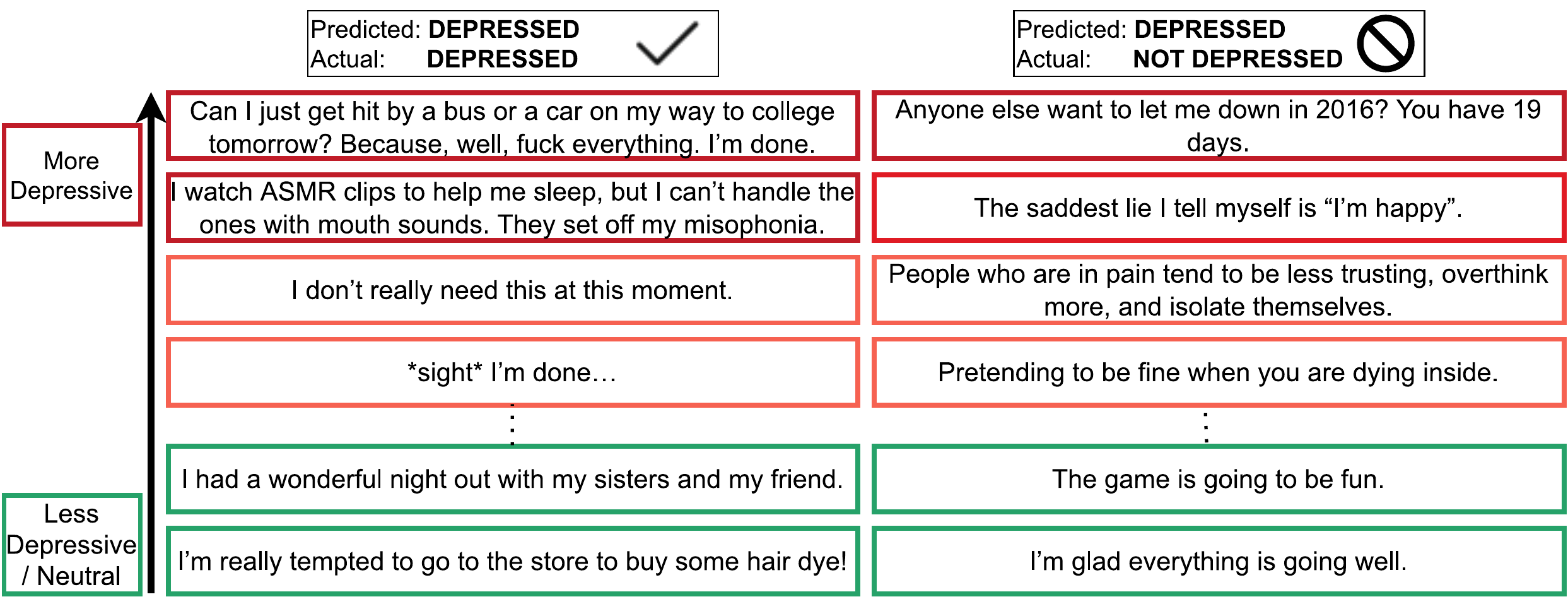}{
    \caption{Error analysis on two predictions, one correct (\textit{Left}), and another incorrect (\textit{Right}). The posts are sorted by their attribution scores given by Integrated Gradients \cite{sundararajan2017axiomatic}. The top posts have strong attribution for a positive (depressed) prediction, the bottom texts have a weak attribution to a positive prediction. \textit{All examples were paraphrased, and only the texts are shown to maintain anonymity.}}
    \label{fig:integrated-gradients}
}
\end{figure}
We perform an error analysis on the predictions of \textit{Time2VecTransformer} on the Twitter data. We use Integrated Gradients \cite{sundararajan2017axiomatic} to extract posts' attributions scores for predictions. In Figure \ref{fig:integrated-gradients} (\textit{Left}), the posts with depression cues have the strongest attribution scores, and the user is correctly labeled by the model. Given the way in which the mental health datasets are annotated by users' mention of diagnosis, some users from the non-depressed class may also have depression, but without mentioning it on social media. This may be the case of the user from Figure \ref{fig:integrated-gradients} (\textit{Right}), who is showing definite signs of sadness and was incorrectly predicted by the model as having depression. Since our model operates across posts, using a feature attribution method such as Integrated Gradients naturally enables the automatic selection of the most relevant posts from a user, similar to \cite{gui2019cooperative} and \cite{rissola2020dataset}, but without relying on a specialized procedure to do so.

\subsection{Limitations and Ethical Considerations}
The method proposed in this paper is trained on data with demographic bias \cite{hovy2016social} from Reddit and Twitter, two social media platforms with a demographic skew towards young males from the United States; thus our method may not succeed in identifying depression from other demographic groups. The aim of our system is to help in detecting cues of depression found in social media, and not to diagnose depression, as the diagnosis should only be made by a health professional following suitable procedures. Further, the dataset annotations rely on the users' self-reports, but without knowing the exact time of diagnosis. Harrigian et al. \cite{harrigian-dredze-2022-now} studied the online content of users with a self-report of depression diagnosis and observed a decrease in linguistic evidence of depression over time that may be due to the users receiving treatment or other factors.



\section{Conclusions}
In this work, we showcased our time-enriched multimodal transformer architecture for depression detection from social media posts. Our model is designed to operate directly at the user-level: the attention mechanisms (both cross-modal and self-attention) attend to text and images across posts, and not to individual tokens. We provided two viable ways to train our method: a time-aware approach, in which we encode the relative time between posts through \textit{time2vec} positional embeddings, and a set-based approach, in which no order is assumed between users' posts. We experimented with multiple sampling methods and positional encodings (i.e., \textit{time2vec}, \textit{zero} and \textit{learned}) and multiple state-of-the-art pretrained text and image encoders. Our proposed \textit{Time2VecTransformer} model achieves state-of-the-art results, obtaining a 0.931 average F1 score on the Twitter depression dataset \cite{gui2019cooperative}. Using the \textit{SetTransformer}, we obtain a 0.902 F1 score on \textit{multiRedditDep} \cite{uban2022explainability}, a multimodal depression dataset with users from Reddit. Given the particularities of the two datasets, we hypothesize that a set-based training regime is better suited to handle datasets containing large amounts of noise and uninformative posts, while a time-aware approach is suitable when user posting frequency is high.

\subsubsection{Acknowledgements} The work of Paolo Rosso was in the framework of the FairTransNLP research project funded by MCIN, Spain (PID2021-124361OB-C31). Liviu P. Dinu was partially supported by a grant on Machine Reading Comprehension from Accenture Labs.

\bibliographystyle{splncs04}
\bibliography{refs}

\end{document}